\documentclass[journal]{IEEEtran}
\usepackage[noadjust]{cite}
\usepackage{amsmath,amssymb,amsfonts}
\usepackage{algorithmic}
\usepackage{subfigure}
\usepackage{graphicx}
\usepackage{textcomp}
\usepackage{xcolor}
\usepackage{soul}
\usepackage{graphics} 
\usepackage{epsfig} 
\usepackage{mathptmx} 
\usepackage{times} 
\usepackage{amsmath} 
\usepackage{amssymb}  
\usepackage{multirow}
\usepackage{graphicx, epstopdf}
\usepackage{xspace,epsfig,url}
\usepackage{psfrag}
\usepackage{verbatim}
\usepackage[all]{xy}
\usepackage{color}
\usepackage{comment}
\usepackage{cases}
\usepackage{hyperref}

\begin{document}
%
\title{Perception of Mechanical Properties via\\ Wrist Haptics: Effects of Feedback Congruence}
%
%
%

\author{Mine Sarac$^{1,2}$,~\IEEEmembership{Member,~IEEE,}, Massimiliano di Luca$^{3}$, and
        Allison M. Okamura$^{2}$,~\IEEEmembership{Fellow,~IEEE,}
\thanks{$^{1}$Kadir Has University, Turkey
        {\tt\footnotesize mine.sarac@khas.edu.tr}}%
\thanks{$^{2}$Stanford University, USA
        {\tt\footnotesize aokamura@stanford.edu}}%
\thanks{$^{3}$University of Birmingham, UK, was at Facebook Reality Labs, USA
        {\tt\footnotesize m.diluca@bham.ac.uk}}%
\thanks{This work was supported in part by National Science Foundation grant 1830163 and Combat Capabilities Development Command-Soldier Center (CCDC-SC) grant W81XWH-20-C-0008.}}

\maketitle

\begin{abstract}
Despite non-co-location, haptic stimulation at the wrist can potentially provide feedback regarding interactions at the fingertips without encumbering the user's hand. Here we investigate how two types of skin deformation at the wrist (normal and shear) relate to the perception of the mechanical properties of virtual objects. We hypothesized that a congruent mapping between force at the fingertips and deformation at the wrist would be better, i.e. mapping finger normal force to skin indentation at the wrist, and shear force to skin shear at the wrist, would result in better perception than other mappings that either mixed or merged the two direction into a single type of feedback. We performed an experiment where haptic devices at the wrist rendered either normal or shear feedback during manipulation of virtual objects with varying stiffness, mass, or friction properties. 
Perception of mechanical properties was more accurate with congruent skin stimulation than noncongruent. 
In addition, discrimination performance and subjective reports were positively influenced by congruence. This study demonstrates that users can perceive mechanical properties via haptic feedback provided at the wrist with a consistent mapping between haptic feedback and interaction forces at the fingertips, regardless of congruence. 
\end{abstract}


%
\IEEEpeerreviewmaketitle

\section{Introduction}

\begin{figure}[t!]
  \centering
  \resizebox{3.4in}{!}{\includegraphics{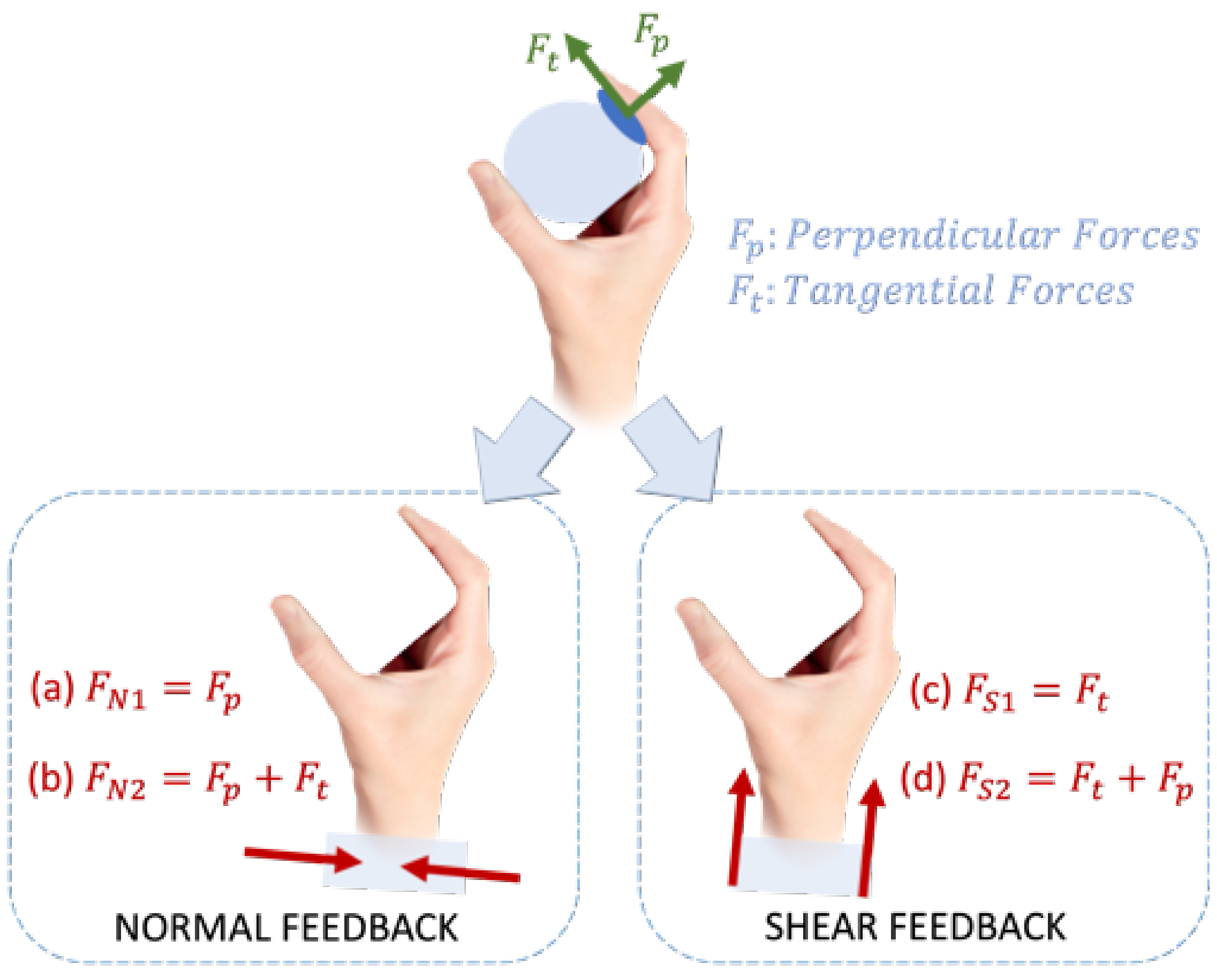}}
  \caption{Haptic conditions used to study perception of mechanical properties: Interaction forces at the virtual fingertip are mapped into relocated haptic feedback cues as (a) only the perpendicular component in the normal feedback direction, (b) both perpendicular and tangential components in the normal direction, (c) only the tangential component in the shear feedback direction, (d) both perpendicular and tangential components in the shear direction.}
  \label{fig:conditions}
  \vspace*{-1.5\baselineskip}
\end{figure}

Mechanical properties of real-world objects, such as mass, stiffness, friction, and temperature are often perceived via direct touch at the fingertip (Fig.~\ref{fig:conditions}). One goal of haptic display is to recreate interaction sensations to make the user perceive these mechanical properties. Many multi-degree-of-freedom fingertip devices have been developed to render interaction forces during active exploration and manipulation tasks in virtual environments~\cite{CulbertsonAR2018,Schorr2017, Leonardis2017}. High level of perceived realism, good dexterity, and useful communication of information (e.g., the mechanical properties of objects) during manipulation tasks can be obtained with high-performance devices having many degrees of freedom.

There are requirements for high-performance haptic devices to reduce encumbrance, complicating design and increasing the cost of actuators. Furthermore, users cannot wear fingertip devices in certain applications, e.g., augmented reality, where it is desirable to leave the fingertips free to interact with physical objects. We consider a different approach to artificial haptic feedback by relocating haptic sensations from the fingertip to the forearm, near the wrist. In doing so, the calculated forces from interactions between the fingertips and manipulated virtual objects are rendered on the skin of the arm.

In this scenario, users cannot receive completely realistic feedback because they interact with virtual objects through their fingers but perceive the haptic feedback on their arms. Here we investigate whether such haptic feedback can create meaningful, believable interactions. 
Such believable haptic feedback should convey useful information about fingertip contact and material properties of objects without increasing cognitive load for the user, such that it qualitatively adds to (rather than detracts from) the user experience. Such relocation has been previously used successfully for social interactions~\cite{Baumann2010, Wang2012, Culbertson2018, Knoop2015, Tsetserukou2010}, communication~\cite{Dunkelberger2018, Song2015, Zheng2012}, navigation~\cite{Nakamura2014}, and teleoperation~\cite{Meli2018}. 

In this study, we aim to examine the effects of direction of force at the wrist on  perception of virtual object mechanical properties (Fig.~\ref{fig:conditions}). Previous wrist-based devices, including Tasbi~\cite{Pezent2019}, Bellowband~\cite{Young2019}, and WRAP~\cite{RaitorICRA2017}, and others \cite{Moriyama2018}, provided haptic feedback to the wrist in a distributed manner and showed that such relocated haptic feedback could create better user perception during virtual interactions tasks compared to no haptic feedback. 

It is unknown how the direction of applied force to a user's skin at or near the wrist should be exploited to enhance the perception of the mechanical properties of virtual objects. We previously  
designed haptic bracelets to render either normal (perpendicular to the skin) or shear (parallel to the skin) feedback near the wrist, and measured users' discrimination of virtual objects based on stiffness with normal vs.\ shear feedback~\cite{Sarac2022}. Our results showed that participants performed better with feedback in the normal direction, and shear feedback did not feel as strong as normal. We then conducted a second study and found that normal and shear stimuli cannot be equalized through skin displacement or the interaction forces across all users. Instead, they should be equalized with a calibration based on the method of adjustments. 
With this information in hand, we now aim to determine how the force directions calculated at the fingertips should be mapped to force directions at the wrist.


\section{Hypotheses and Experiment Design}

Interaction forces are computed between a user's finger avatars and a manipulated object, both perpendicular and tangential to the fingertip surface. 
While discriminating stiffness, interaction forces occur perpendicular to the fingertips as the participant squeezes the object~\cite{Lederman2011, Klatzky1987}. While discriminating mass or friction, interaction forces occur tangential to the fingertips as the participant lifts object between two fingertips (for mass) and slides the index finger tangentially to the surface (for friction). 
We designed a study to 
investigate the impact of congruent vs.\ noncongruent mappings between fingertip forces and the deformation at the wrist using haptic conditions shown in Fig.~\ref{fig:conditions}. 


We hypothesized that participants' perception of mechanical properties would be better with a congruent mapping between force at the fingertips and deformation at the wrist than merged or noncongruent mappings. The advantages of such congruence are clear for rendering at the fingertips \cite{CulbertsonAR2018,Pacchierotti2017}, and we propose that this should extend to relocated haptic feedback. In addition, due to limited perception at the wrist compared to the fingertip \cite{Johansson2009}, we hypothesized that simplifying feedback to a single direction will result in more meaningful, interpretable feedback. Our experiment uses two different sets of tasks to investigate these hypotheses.

\begin{itemize}
\item In one set (main discrimination trials), we measure how well participants can discriminate mechanical properties (stiffness, mass, and friction). Here we inform participants which mechanical property varies for a given task, and instruct them on how to explore the virtual objects while experiencing different haptic conditions (normal vs./ shear haptic feedback). This allows us to measure the participants' just noticeable difference for each mechanical property. 
\item In another set of tasks (open-response trials), we present participants with a pair of virtual objects \emph{without} telling them which mechanical property varies or how to interact with the virtual objects. This allows us to determine how participants interpret the rendered object mechanical properties. These pre- and post- studies were performed at the beginning and end of each main discrimination block. 
\end{itemize}

\section{Experiment Setup}

\begin{figure}[t!]
  \centering
  \resizebox{3.4in}{!}{\includegraphics{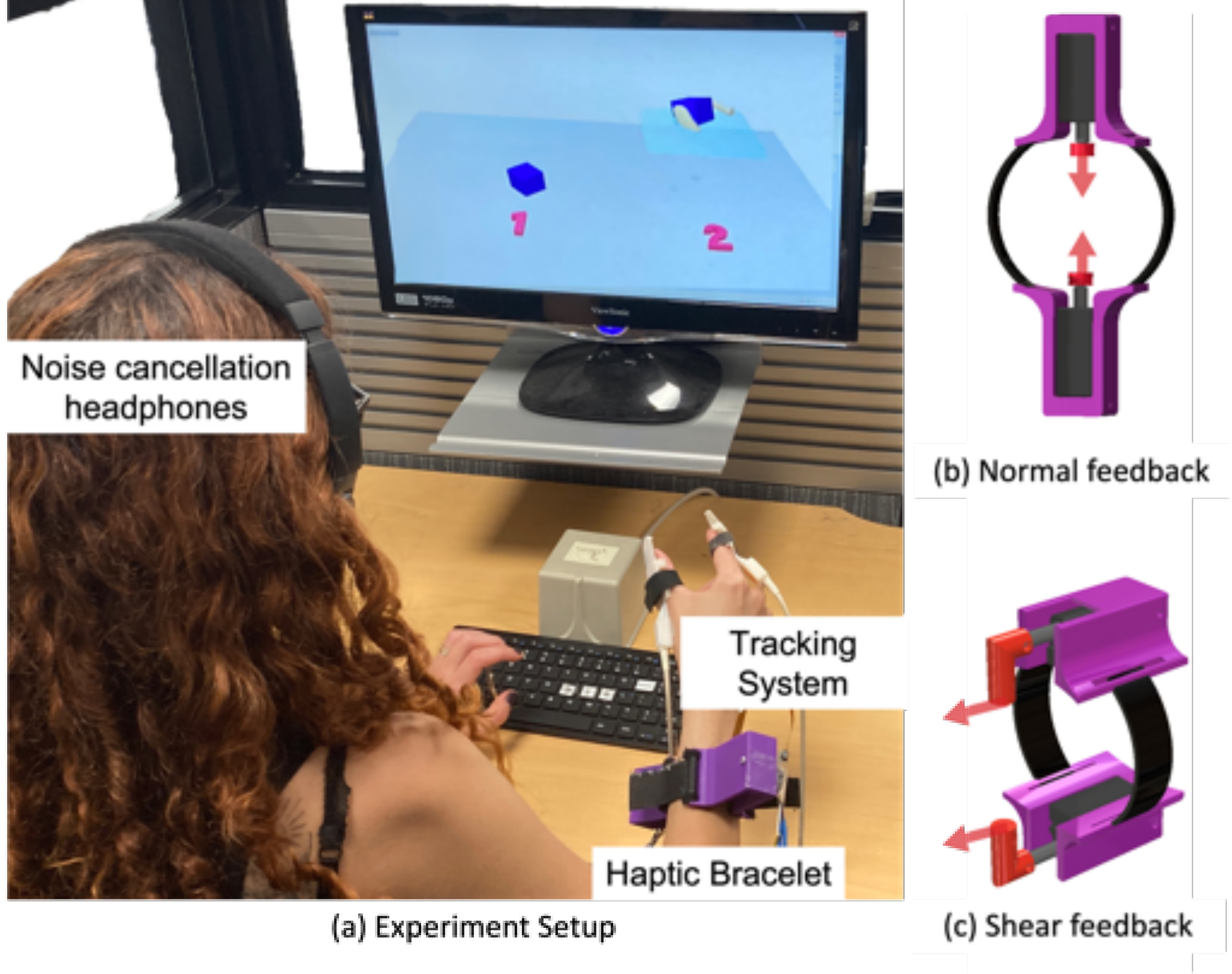}}
  \vspace*{-1\baselineskip}
  \caption{Experiment setup adapted from our previous work \cite{Sarac2022}: (a) The participant sits in front of a monitor and wears a fingertip tracking sensor and noise cancellation headphones. She interacts with objects in the virtual environment while interaction forces are rendered on the wrist through haptic bracelets in the (b) normal direction or (c) shear direction.}
  \label{fig:setup}
  \vspace*{-.5\baselineskip}
\end{figure}

Fig.~\ref{fig:setup} shows the experiment setup. 
The participant sits interacts with objects in a 3D virtual environment where a table, an object, and two fingertips are visible. An electromagnetic tracking system tracks movements of the fingertips in real time, while haptic bracelets render the virtual interaction forces on the wrist. To minimize the effect of environment and actuator auditory noise, the participant wears headphones that play white noise and cancel external noise.

\begin{figure*}[t!]
  \centering
  \resizebox{6.5in}{!}{\includegraphics{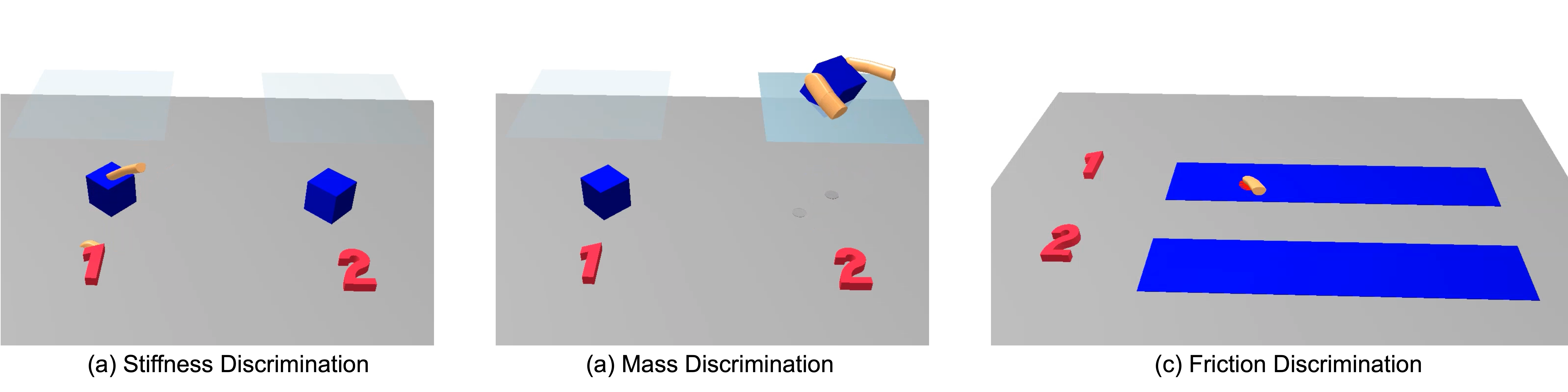}}
  \vspace*{-1\baselineskip}
  \caption{Experiment tasks in the virtual reality environment for (a) stiffness discrimination, (b) mass discrimination, and (c) friction discrimination.}
    \vspace*{-1\baselineskip}
  \label{fig:task}
\end{figure*}

\subsection{Wrist Haptic Devices (Haptic Bracelets)}

Each haptic bracelet comprises two actuator sets on the dorsal and ventral sides of the forearm and weighs less than 40 g. The orientation of the actuator results in normal or shear forces on the user's skin, as shown in Fig.~\ref{fig:setup}. Each bracelet uses an Actuonix PQ12-P linear actuator due to its low weight (15~g), maximum stroke (20~mm), high output force (18~N), and straightforward control using an integrated position sensor. 
Users wear the bracelet on their forearm near the wrist to minimize the impact of wrist movements and facilitate consistent physical contact. The grounding is designed with a curvature to fit the forearm, a silicone pad between the plastic and the skin, and wide Velcro straps to keep the grounding stable. Further details about the device design and performance are described in~\cite{Sarac2022}.

\subsection{Virtual Environment}

We created a virtual environment using the CHAI3D framework~\cite{Conti2005} as shown in Fig.~\ref{fig:setup}. The virtual environment is displayed on a 2D monitor and updated at 144~Hz. The user's finger movements are tracked at approximately 200~Hz using a trakSTAR tracking system and an Ascension Model 800 sensor attached to the user's finger through 3D printed grounding. The virtual environment displays the virtual finger pose, as well as the objects in the virtual environment with which the user interacts (Fig.~\ref{fig:task}).

\subsection{Participants}
14 participants (age 23-33, 7 females and 7 males) joined the study. All participants were right-handed. The Stanford University Institutional Review Board approved the experimental protocol, and all participants gave informed consent. Before the experiment, participants reported their haptic experience on a scale between L0 (no experience) and L7 (expert). 1 participant reported L2, 3 participants reported L3, 2 participants reported L5, 1 participant reported L6, and 7 participants reported L7. 

\section{Experiment Procedures}

\subsection{Calibration}

To compare the perceptual and performance differences of normal and shear feedback rendered on the wrist, we must ensure that the participants perceive both stimuli equally~\cite{Sarac2022}. Upon arrival, the experimenter tightened the bracelets (first on the right side, then on the left) using measurement marks assigned on the Velcro straps. Both bracelets contacted the skin 35 mm away from the wrist bone. The tightness is adjusted until the participant confirms that both bracelets feel equally intense with no actuation, then the following calibration using the method of adjustments is performed: The participant is given 1 and 3 mm reference displacements of the bracelet actuators in the normal direction, and asked to adjust the displacement in the shear directions, as in \cite{Sarac2022}. For each reference value, a staircase method was used to find the shear displacement that created a sensation of equal intensity to the normal reference. Using the adjusted displacements for both reference values, we modeled a linear, personalized relationship between normal and shear feedback for each participant.

On day 1, we recorded the measurement marks assigned on the Velcro straps for each participant. These recordings were used to tighten the Velcro straps with the same intensity as day 1 - so no additional calibration was needed. 

\subsection{Main Discrimination Trials}
\label{sec:main}

The participants see two identical virtual objects with different simulated mechanical properties (Fig. \ref{fig:task}) and move their fingers to interact with the objects. In the main discrimination trials, participants performed a two-alternative forced-choice task in which they explored the virtual objects and then report which object had the larger value of the mechanical property relevant to that task (stiffness, mass, or friction). 
The procedure for each mechanical property was as follows:

\begin{figure*}[t!]
  \centering
  \resizebox{7in}{!}{\includegraphics{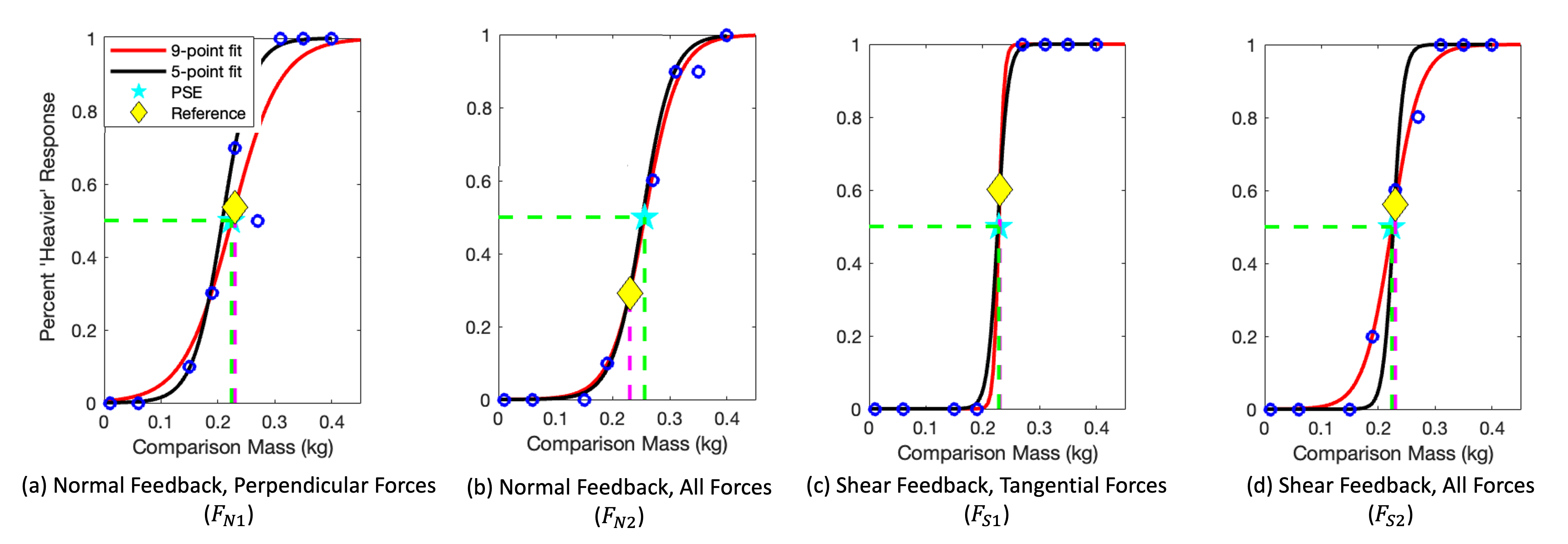}}
  \vspace*{-.5\baselineskip}
  \caption{Sample psychometric curves during mass discrimination for each of the four haptic conditions with 9 data points (red), and 5 data points selected among the recorded 9 points (black). The psychometric curves were fit to the participant’s responses and used to calculate the PSE and JND for each haptic condition.}
  \label{fig:curve}
\end{figure*}

 \textbf{Stiffness:} In each trial, participants were presented with two identical-looking cubes with different stiffness values and identical mass and friction values (Fig. \ref{fig:task}(a)). The participant squeezed each object to evaluate the stiffness, either by pushing down with one finger from the top, squeezing with two fingers from the sides when the object is on the ground, or squeezing with two fingers from the sides when the object is lifted. Participants could choose one of these strategies, change their strategy when desired, or use alternative strategies while evaluating a pair of objects. Participants were allowed to interact with each object as many times as desired until they were ready to report which object felt stiffer.
 
 \textbf{Mass:} In each trial, participants were presented with two identical-looking cubes with different mass values and identical stiffness and friction values (Fig. \ref{fig:task}(b)). The participant grabbed each object using two fingers (thumb and index finger) and lifted the object until reaching a target plane located near the ceiling of the work space. As the object passed through the target, its opacity changed from semi-transparent to completely opaque. This visual cue indicated to the participant that they lifted the object a sufficient distance to have explored its mass. Participants were allowed to interact with each object as many times as desired until they were ready to report which object felt heavier. Participants were allowed to give an answer only if both targets had become opaque. 
 
\textbf{Friction:} In each trial, participants were presented with two identical-looking strips on the ground of the virtual environment with different friction values and identical stiffness and mass values (Fig. \ref{fig:task}(c)). Participants pushed their index finger onto the strip until the shadow of the index finger turned red, and then slid their finger along the strip. The visual cue guided the participant to push on the strip a sufficient amount to generate friction force. Participants were allowed to interact with each strip as many times as desired until they were ready to report which object seemed to have more friction. 

To choose the stiffer/heavier/higher-friction objects, participants typed on the keyboard the number that appeared next to the corresponding virtual object. Participants then pressed on another key to start the next trial.



\subsection{Open Response Trials}
\label{sec:open}

In each trial, participants were presented with two identical-looking cubes in the virtual environment, similar to the mass trials shown in Fig~\ref{fig:task}(a). The cubes differed in one mechanical property and were the same in the other two mechanical properties. Unlike the main discrimination trials, participants were not told which property differed between the two cubes. Rather, they were asked to comment verbally on any difference they felt. Participants were allowed to interact with each object as many times as desired until they provided a response.

\subsection{Experiment Flow}

\begin{figure}[t!]
  \centering
  \resizebox{3in}{!}{\includegraphics{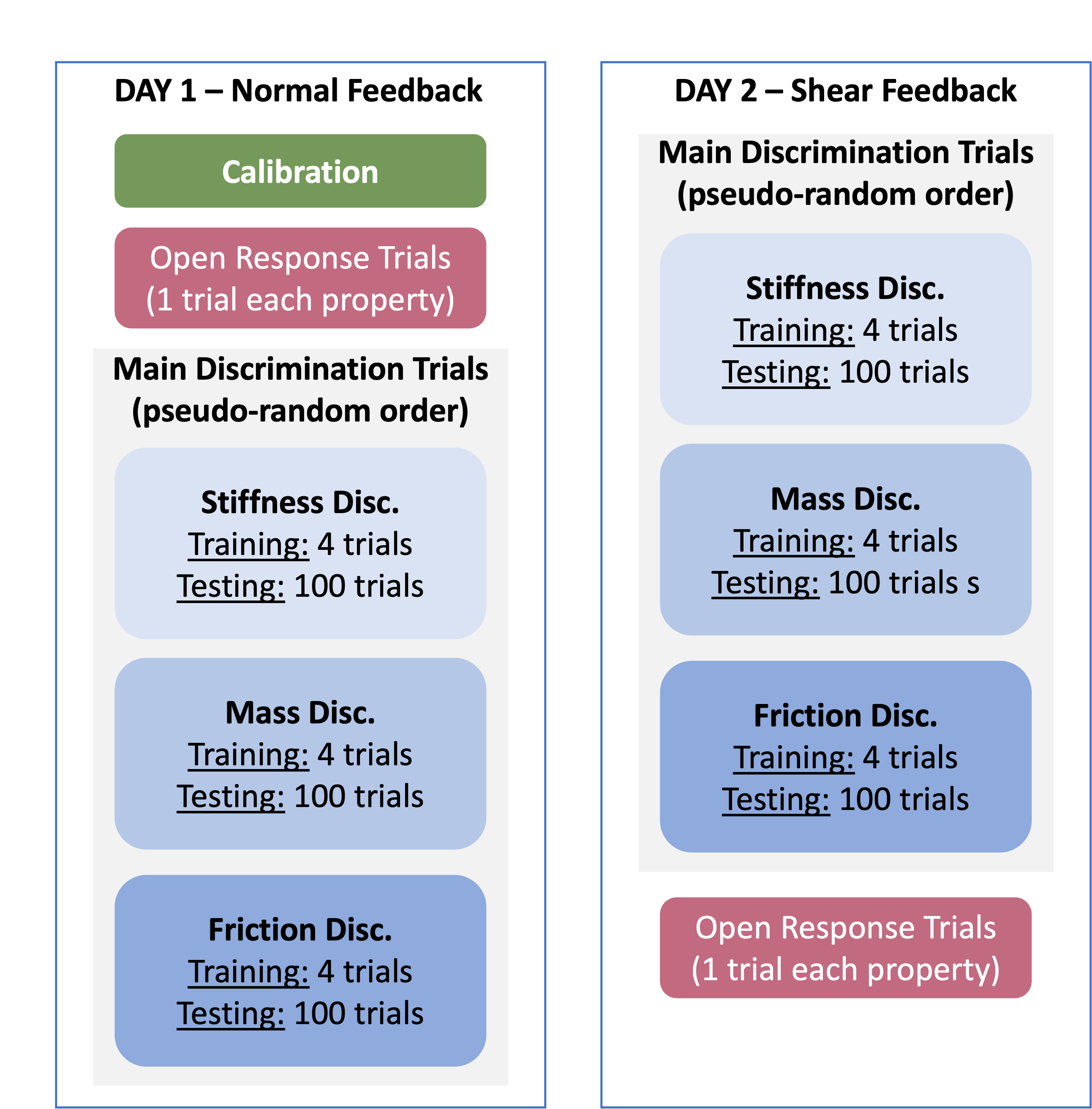}}
  \caption{Experiment flow and the number of trials in terms of the feedback direction, experiment sections, and mechanical properties to discriminate. The comparison values, the order for the stiffness/mass/friction discrimination blocks, and feedback direction assigned to each day were pseudo-randomized for each participant.}
    \vspace*{-0.5\baselineskip}
  \label{fig:flow}
\end{figure}

Combining the two types of trials described above, we designed the overall experiment as in Fig. \ref{fig:flow}. The trials were conducted over two days with a maximum of 2 hours per day. The order for the normal/shear feedback and the stiffness/mass/friction discrimination blocks within the main experiment were pseudo-randomized across participants. There were 2-minute breaks after every 50 trials and 5-minute breaks between each mechanical property block.

\subsection{Metrics and Analysis}

Participants’ responses were recorded during the experiment and analyzed based on subjective comments, time spent on each trial, and the discrimination performance. Subjective comments are reported to help us understand the overall experience and participants’ preferences. Time spent indicates the complexity of the given discrimination task or different comparison pairs. Discrimination performance is analyzed by averaging the correct answers while identifying the object with the higher value. The average values obtained from different comparison pairs can create a psychometric curve using:
\begin{equation} \label{eqn}
   y = \frac{1}{1+e^\frac{\alpha-x}{\beta},} 
\end{equation}
\noindent where y is the proportion of “heavier”, “stiffer”, or “rougher” responses, $x$ is the comparison value, $\alpha$ is the point of subjective equality (PSE), and $\beta$ is a slope fitting parameter. The values of $\alpha$ and $\beta$ are determined from the fit of the sigmoid function. Just noticeable difference (JND) is then calculated by subtracting the PSE ($\alpha$) from the comparison weight corresponding to $y = 0.75$. Fig.~\ref{fig:curve} shows sample psychometric curves fit a single participant’s responses collected during mass discrimination for all haptic conditions. 

\begin{figure*}[t!]
  \centering
  \vspace*{-.5\baselineskip}
  \resizebox{7.4in}{!}{\includegraphics{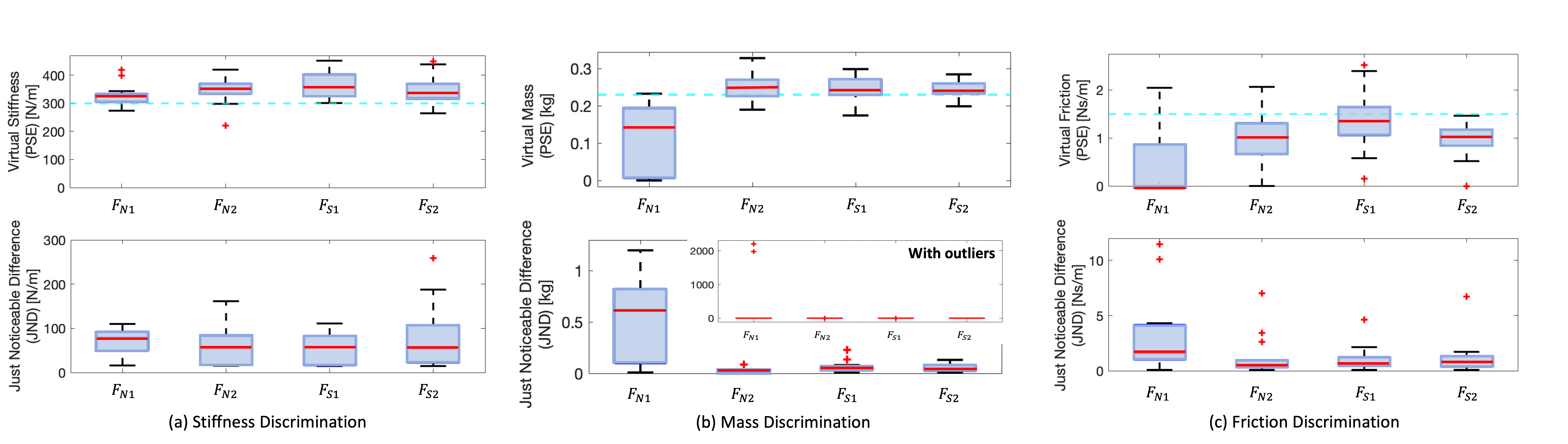}}
  \vspace*{-1\baselineskip}
  \caption{Box-and-whisker plots of PSEs and JNDs for each haptic condition across all 14 participants during (a) stiffness discrimination, (b) mass discrimination, and (c) friction discrimination task. Haptic condition $F_{N1}$ is normal feedback with perpendicular forces, $F_{N2}$ is normal feedback with all forces, $F_{S1}$ is shear feedback with tangential forces, and $F_{S2}$ is shear feedback with all forces. Median values for JND and PSE are reported in Tables \ref{fig:averages1} and \ref{fig:averages2}.}
  \label{fig:results}
\end{figure*}

\subsection{Pilot Study and Reference/Comparison Values}

We performed a pilot study to identify appropriate values of reference and comparison mechanical property values to measure JNDs. A single user performed ten trials for each of nine comparison values, three mechanical properties, and four haptic conditions (see Fig.~\ref{fig:conditions}). The orders of the comparison values here and in the main study were pseudo-randomized within the blocks of fixed varying property to minimize effects of presentation order. Through this process, we found a set of comparison values that resulted valid psychometric curves for all haptic conditions. Five comparison values were deemed sufficient to reproduce the psychometric curve with a reasonable-length experiment to avoid user fatigue.

As a result of the pilot study, we defined the reference and comparison values for each mechanical property during the main discrimination trials of the experiment described in Section~\ref{sec:main}. All trials used a reference stiffness of 340 N/m and comparison stiffness values of 80, 210, 340, 470, and 600 N/m for the stiffness discrimination task. All trials used a reference mass of 0.23 kg and comparison mass values of 0.06, 0.15, 0.23, 0.31, and 0.4 kg for the mass discrimination task. Finally, all trials used a reference friction of 1.5 Ns/m and comparison friction values of 0.1, 0.75, 1.5, 2.25, and 3 Ns/m for the friction discrimination task. 

For the open-response trials described in Section~\label{sec:open}, participants interacted with two objects with the minimum and the maximum comparison values from the discrimination trials (80 to 600 N/m for stiffness, 0.23 to 0.4 kg for mass, and 0.1 to 3 Ns/m for friction). 

\section{Results}

All 14 participants completed the experiment. The impact of feedback congruence were investigated through the statistical analysis performed on PSE and JND values calculated through users' responses during main discrimination trials. We found no impact of feedback congruence on JNDs, which is related to how well participants respond to changes with the comparison pairs while discriminating all mechanical properties. However, we found a significantly negative impact of feedback noncongruence on PSE, which is related to the perceived mechanical property while discriminating mass and friction. In addition, the perceived mechanical property were found to be \textit{not} significantly different than the intended reference \textit{only} with feedback congruence. 

We will also report the questionnaire results filled by the participants after the experiment. There is no consensus on the experiment difficulty or the favorite feedback direction. However most participants verbally commented that "haptic feedback did not feel responsive" to their interactions in the virtual environment. Even though they commented on the negative impact of haptic congruence, they did not seem to realize the feedback congruence - compared to the conditions where all fingertip forces are rendered.

\subsection{Main Discrimination Trials}

Fig. \ref{fig:results} shows box-and-whisker plots for PSE and JND across all participants and haptic conditions for discrimination of stiffness, mass, and friction. Tables \ref{fig:averages1} and \ref{fig:averages2} summarize the results. 

\begin{table}[t!]
  \centering
    \caption{Mean and median PSE for all haptic conditions\\ across 14 participants} 
  \vspace*{-.75\baselineskip}
  \resizebox{3in}{!}{\includegraphics{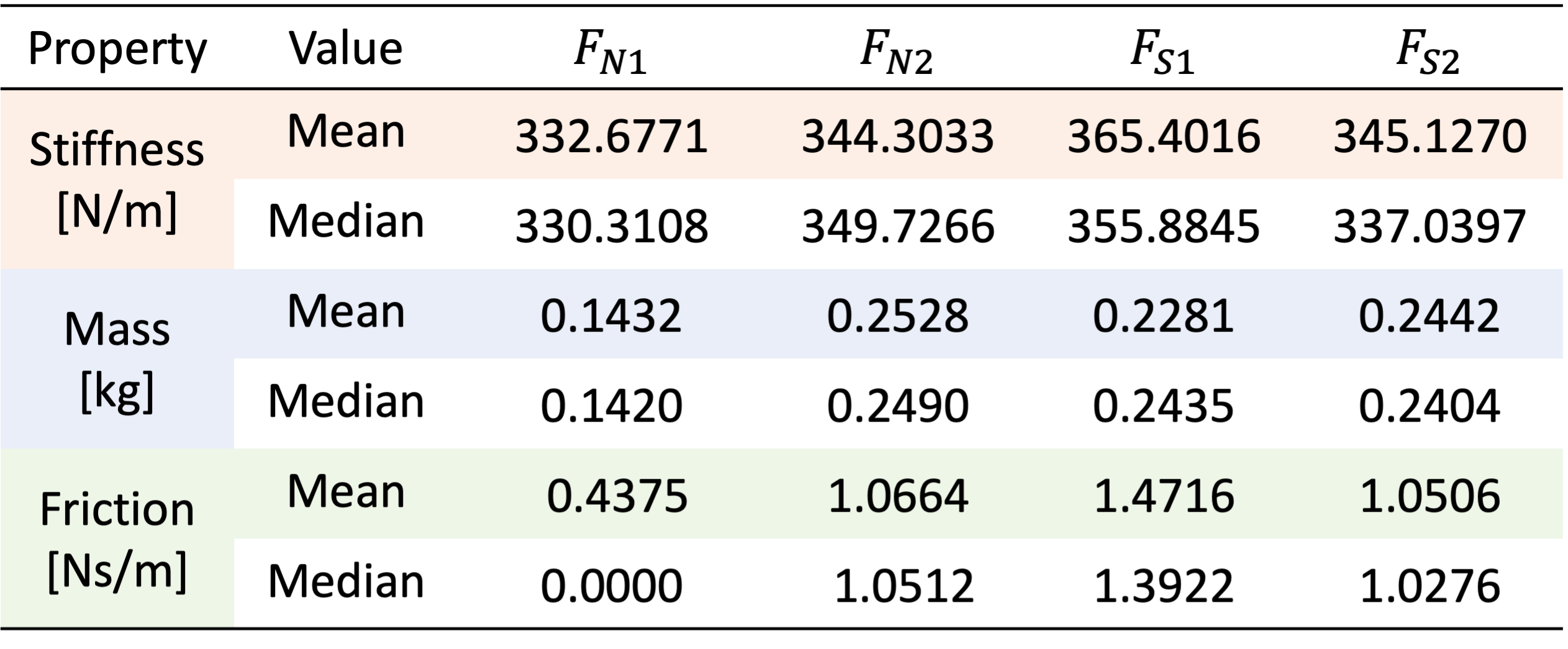}}
    \vspace*{-1.5\baselineskip}
  \label{fig:averages1}
\end{table}

\begin{table}[t!]
  \centering
    \caption{Mean and median JND for all haptic conditions\\ across 14 participants} 
  \vspace*{-.75\baselineskip}
  \resizebox{3in}{!}{\includegraphics{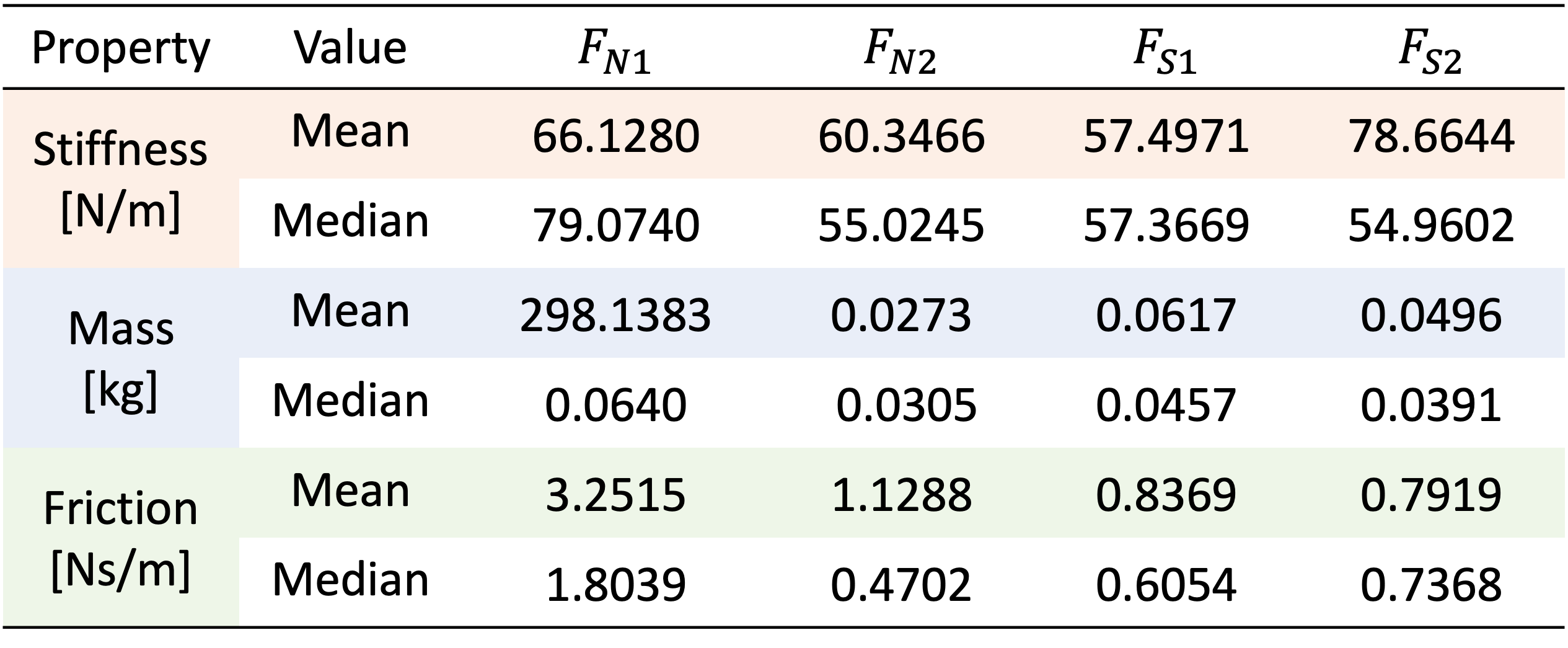}}
    \vspace*{-1.5\baselineskip}
  \label{fig:averages2}
\end{table}

\subsubsection{Stiffness Discrimination} 

We performed a one-way repeated-measures analysis of variance (ANOVA) on PSE and JND with haptic condition as the main factor. There was no statistically significant effect of haptic condition on PSE ($F(3,55) = 1.120, p=0.349, \eta^2_{partial} = 0.061$) or JND ($F(3,55) = 0.537, p = 0.659, \eta^2_{partial} = 0.030$). We also performed independent t-tests to determine if there was a significant difference between PSE and the reference stiffness value for each haptic condition. The results showed that PSE was significantly different from the reference \textit{only} with noncongruent feedback ($F_{S1}$, $p=0.01367$). 


The lack of difference between haptic conditions might be because of the simplicity of the stiffness discrimination task. Even though participants were allowed to choose how to investigate the stiffness properties, the experimenter observed that most participants chose to squeeze the objects with one or two fingers without lifting them. Thus, the interaction forces were not influenced by mass and friction. Most participants also reported the stiffness discrimination task to be the easiest, which might have allowed them to perform well under all haptic conditions. 

\subsubsection{Mass Discrimination} 


We performed a one-way repeated-measure ANOVA on PSE and JND with haptic condition as the main factor. There was no statistically significant effect of haptic condition on 
JND ($F(3,55) = 2.165, p = 0.103$) but we found a statistical significance on PSE ($F(3,55) = 19.127, p < 0.001$). A post-hoc Tukey test showed that the haptic condition of normal direction with perpendicular forces ($F_{N1}$) was statistically significantly different than all other conditions ($p<0.001$), while there was no difference between the rest of the conditions. 

We also performed independent t-tests to determine if there was a significant difference between PSE and reference mass value for each haptic condition. The results showed that PSEs are significantly different than reference when all interaction forces are rendered in the normal direction ($F_{N2}$, $p=0.0487$) and in the shear direction ($F_{S2}$, $p=0.0377$); but not with feedback congruence ($F_{S1}$, $p=0.0749$).

\subsubsection{Friction Discrimination} 


We performed a one-way repeated-measures ANOVA on PSE and JND with haptic condition as the main factor. There was no statistically significant effect of haptic condition 
on JND ($F(3,55) = 2.483, p = 0.071$) but we found a statistical  significance on PSE ($F(3,55) = 7.151, p < 0.001$). A post-hoc Tukey test showed that the haptic condition of normal direction with perpendicular forces ($F_{N1}$) was statistically significantly different than all other conditions ($p<0.001$), while there was no difference among the rest.

We also performed independent t-tests to determine if there was any significant difference between PSE and the reference friction value for each haptic condition. The results showed that PSEs are significantly different than the reference when all interaction forces are rendered in the normal direction ($F_{N2}$, $p=0.0032$) and in the shear direction ($F_{S2}$, $p<0.001$), but not with feedback congruence ($F_{S1}$, $p=0.8568$).

\subsection{Open Response Results}

Before and after the main discrimination trials, participants were asked to identify any mechanical differences between two identical-looking cubes, where one mechanical property, unbeknownst to the participants, differed. Participants were asked to verbally state what felt different between the two cubes in their own words. Table \ref{fig:openquest} shows the number of participants who identified each varying property based on the experimenter's interpretation of the participants' verbal responses before (pre-study) and after (post-study) the main discrimination trials.
\begin{table}[t!]
  \centering
    \caption{Results of open-response trials to identify differences in mechanical properties between two objects, as perceived by haptic feedback applied to participants’ wrists.}
  \resizebox{3.4in}{!}{\includegraphics{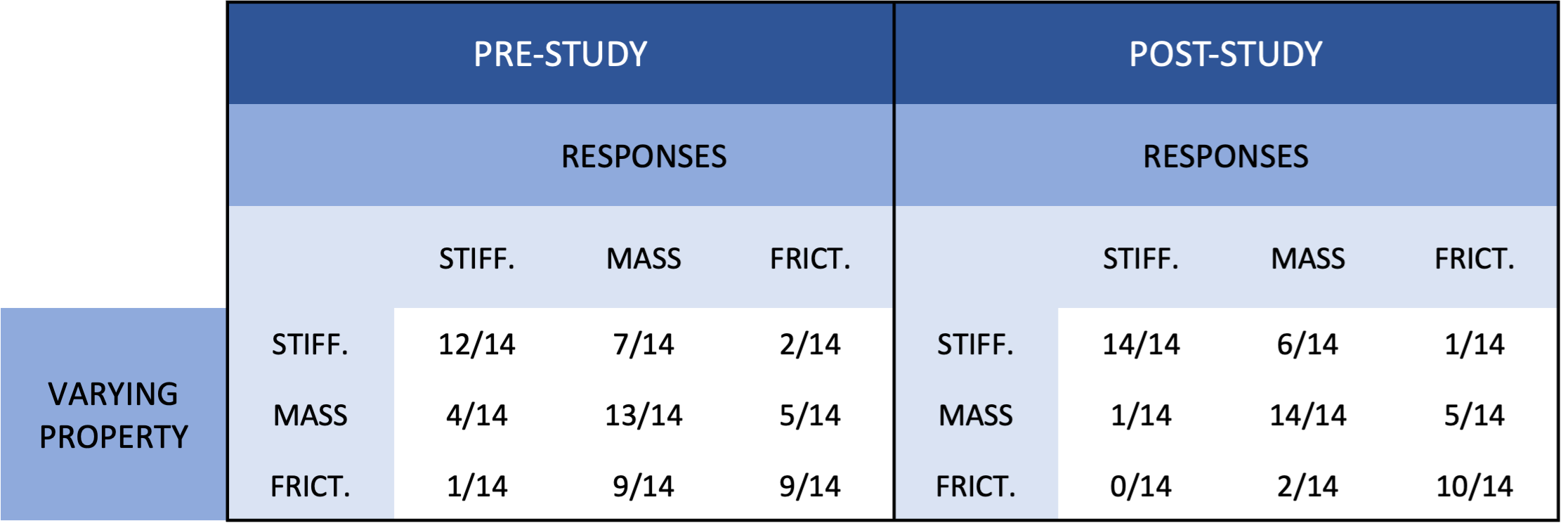}}
    \vspace*{-1.5\baselineskip}
  \label{fig:openquest}
\end{table}

Previously, a similar experiment was performed with six participants using fingertip (rather than wrist) haptic devices \cite{Schorr2017}. Participants identified friction better with the fingertip devices than with the haptic bracelet, but they identified stiffness better with the haptic bracelet, and they identified mass similarly. This indicates that perception of mechanical properties with the haptic bracelets is possible, and could be similar in performance to fingertip haptic devices.

\subsection{Subjective Comments}

Each participant completed a survey about the experiment. 

\textbf{Q1 - Experiment Difficulty:} We asked participants to evaluate the difficulty of the overall experiment on a scale from L1 (easy) to L7 (difficult). One participant reported L1, 3 participants reported L2, 3 participants reported L3, 4 participants reported L4, 2 participants reported L5, and 1 participant reported L6. We observed an inverse relationship between the participants’ responses on the self-evaluation scale and the experiment difficulty scale. The participant who reported L2 on the haptic experience scale reported L6 on the difficulty scale. In contrast, the participants who reported L7 on the haptic experience scale chose either L1, L2, or L3 on the difficulty scale. 

\textbf{Q2 - Task Difficulty:} We asked participants to compare the difficulty of the discrimination tasks. The participants graded stiffness, mass, and friction on a scale of 1 (easiest) to 3 (hardest). Fig. \ref{fig:difficulty} shows the task difficulty scale across participants, and the numbers shown inside each square indicate the number of participants reporting the corresponding answer. Before the experiment, we expected that stiffness would be easiest because the exploration involved squeezing without sliding the finger through the object or lifting the object. Thus, the interaction force is calculated solely based on the stiffness information. Such simplicity might be perceived as easiness of the discrimination task. We also expected that friction would be the most difficult because the rendered forces directly depend on the velocity of the user’s finger movement, causing the task to be more dynamic than the rest.

\begin{table}[h!]
  \centering
  \caption{Self-assessment task difficulty scale for\\ stiffness, mass, and friction discrimination tasks}
\vspace*{-.75\baselineskip}
  \resizebox{2.6in}{!}{\includegraphics{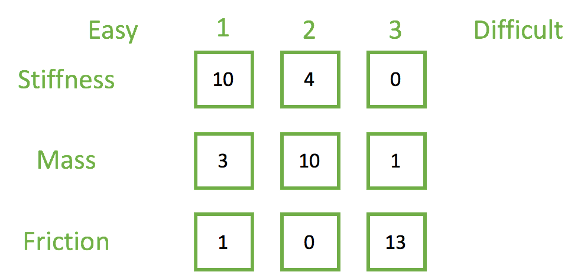}}
  \label{fig:difficulty}
\end{table}

Most participants reported that the stiffness was the easiest among the three discrimination tasks, and the friction was the hardest among the three tasks, as expected. 

\textbf{Q3 - Feedback Direction:} We asked participants also to report which feedback direction they enjoyed the most. 8 participants reported shear, 5 participants reported normal, and 1 participant reported the same. In addition, we asked them to report which feedback direction was easier to notice during the discrimination experiment. 5 participants reported shear, 4 reported normal, and 5 reported the same. Most participants commented that they enjoyed experiencing shear feedback over normal, even though we found no difference in measured JND between normal and shear. 

\section{Conclusion}

In this work, we compared the effects of haptic feedback rendered on users' wrists in the normal and shear directions with alternative force mapping modalities while discriminating stiffness, mass, and friction values of virtual objects. Our results showed that participants’ perception of mechanical properties was not affected by the feedback direction. However, their perception of mechanical properties was better when they received congruent feedback, i.e., the relevant single interaction force component from the fingertip was mapped to the wrist, compared to noncongruent feedback, i.e. irrelevant single force components or multiple force components. 
The subjective comments we collected showed no consensus on preferred feedback directions acting on the skin. While there might be various reasons why designers should choose one feedback direction or another, user performance in identifying mechanical properties of virtual objects is not significantly affected by feedback direction.  

In the future, we will investigate the effects of congruent and noncongruent force mappings under more realistic use cases. We will also study the perceptual differences between relocating the haptic feedback to the wrist and rendering the haptic feedback on the fingertips. In addition, we will extend the work to haptic stimulation at other locations on the body, such as the upper arm and waist, which might be advantageous for various applications, including navigation and social touch.

\section*{Acknowledgment}
The authors thank collaborators at Facebook Reality Labs and Triton Systems, Inc.\ for their guidance on this work.

\ifCLASSOPTIONcaptionsoff
  \newpage
\fi



%
\bibliographystyle{IEEEtran}
\bibliography{HapticBracelet_ICRAwRAL}

%








\end{document}